\pdfoutput=1

\documentclass[11pt]{article}

\usepackage{acl}

\usepackage{times}
\usepackage{latexsym}

\usepackage[T1]{fontenc}

\usepackage[utf8]{inputenc}

\usepackage{microtype}

\usepackage{color}
\definecolor{lightgray}{RGB}{211,211,211}
\usepackage{amsmath}
\usepackage{amssymb}
\usepackage{amsthm}
\usepackage{enumitem}
\setlist{nolistsep}
\usepackage{graphicx}
\usepackage{booktabs}  
\usepackage{threeparttable}  
\usepackage{multicol}  
\usepackage{multirow}  
\usepackage{mathrsfs}

\def\eg{\emph{e.g.}}

\definecolor{good}{RGB}{48,128,20}
\newcommand{\good}[1]{\textcolor{good}{\textbf{#1}}}
\newcommand{\bad}[1]{\textcolor{red}{\textbf{#1}}}

\title{{S$^2$SQL}: Injecting Syntax to Question-Schema Interaction Graph Encoder \\ for Text-to-SQL Parsers}
\author{Binyuan Hui \textsuperscript{\rm 1}, Ruiying Geng \textsuperscript{\rm 1}, Lihan Wang \textsuperscript{\rm 1,2,3 \thanks{\quad Work done during an internship at Alibaba DAMO.}}, Bowen Qin \textsuperscript{\rm 1,2,3 \footnotemark[1]}, Yanyang Li \textsuperscript{\rm 4}, \\ {\bf Bowen Li \textsuperscript{\rm 1}, Jian Sun \textsuperscript{\rm 1} and Yongbin Li \textsuperscript{\rm 1 \thanks{\quad Corresponding author.}} } \\
\textsuperscript{\rm 1} Alibaba Group, \textsuperscript{\rm 2} University of Chinese Academy of Sciences \\
\textsuperscript{\rm 3} Shenzhen Institutes of Advanced Technology, Chinese Academy of Sciences \\
\textsuperscript{\rm 4} The Chinese University of Hong Kong \\
{\small \texttt{\{binyuan.hby,ruiying.gry,yanjin.lbw,jian.sun,shuide.lyb\}@alibaba-inc.com} } \\
}

\begin{document}
\newpage
\maketitle
\begin{abstract}
The task of converting a natural language question into an executable SQL query, known as text-to-SQL, is an important branch of semantic parsing. The state-of-the-art graph-based encoder has been successfully used in this task but does not model the question syntax well. In this paper, we propose \textbf{S$^2$SQL}, injecting \textbf{S}yntax to question-\textbf{S}chema graph encoder for Text-to-\textbf{SQL} parsers, which effectively leverages the syntactic dependency information of questions in text-to-SQL to improve the performance. We also employ the decoupling constraint to induce diverse relational edge embedding, which further improves the network's performance. Experiments on the Spider and robustness setting Spider-Syn demonstrate that the proposed approach outperforms all existing methods when pre-training models are used, resulting in a performance ranks first on the Spider leaderboard.
\end{abstract}

\section{Introduction}
Relational databases are ubiquitous and store a great amount of structured information. The interaction with databases often requires expertise on writing structured code like SQL, which is not friendly for users who are not proficient in query languages. Text-to-SQL aims to automatically translate natural language questions into executable SQL statements~\cite{zelle1996learning,ZettlemoyerC05,WongM07,zettlemoyer2007online,berant2013semantic,Li2014ConstructingAI,Yaghmazadeh2017SQLizerQS,iyer2017learning}.

Recently, a large-scale, multi-table, realistic text-to-SQL benchmark, Spider \cite{YuZYYWLMLYRZR18}, has been released.
The most effective and popular encoder architecture on Spider is the question-schema interaction graph \cite{WangSLPR20}. Built on that, many state-of-the-art models have been further developed \cite{Chen2021ShadowGNNGP, lgesql}.
It jointly models natural language question and structured database schema information, and uses some pre-defined relationships to carve out the interaction between them.
However, we observed that the current graph-based model yet has two major limitations.

\begin{figure}
    \small
    \centering
    \includegraphics[width=0.49\textwidth]{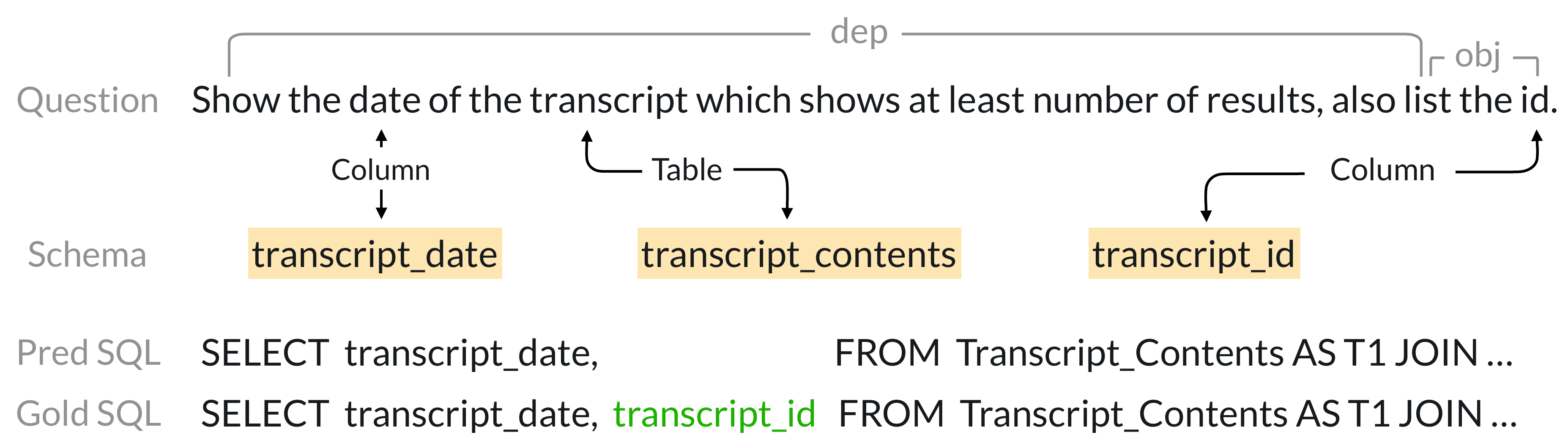}
    \caption{A typical bad case of the current graph-based methods. If the structure of question (dependencies) are not considered, the wrong SQL will be generated even if the linking is correct.}
    \label{intro}
\end{figure}

\begin{figure*}
    \small
    \centering
    \includegraphics[width=0.85\textwidth]{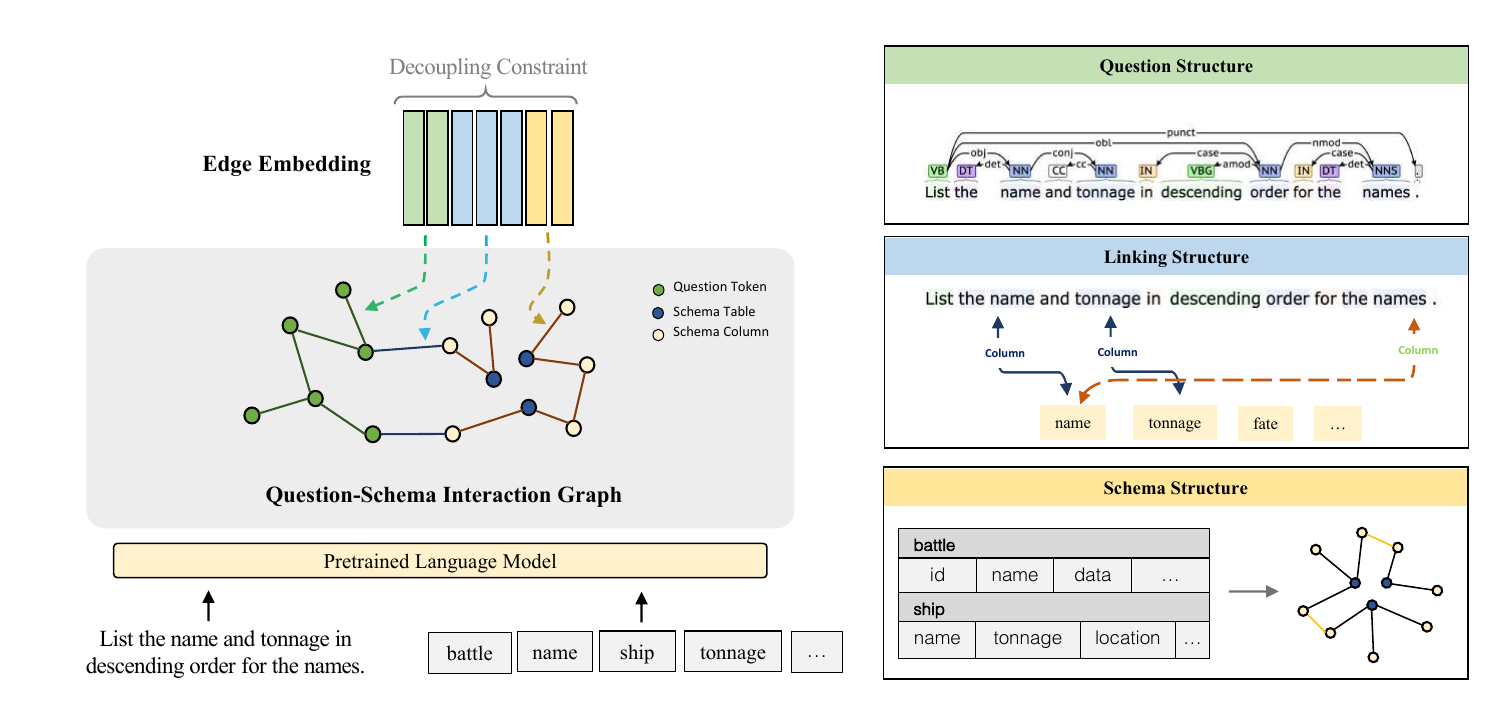}
    \caption{An overview of S$^2$SQL framework. The S$^2$SQL has three relation types to represent the known syntactic information, linking structure and schema information. There structure are integrated into the question-schema interaction graph by learnable edge embedding with decoupling constraints.}
    \label{fig2}
\end{figure*}

\paragraph{\textit{\textbf{Syntactic Modelling.}}}
Jointly modeling syntax and semantics is a core problem in NLP.
In the paradigm of deep learning, the role of syntax should be better understood for tasks in which syntax is a central feature \cite{Ge2005ASS,Michalon2016DeeperSF, Zhang2019SyntaxEnhancedSS,Zanzotto2020KERMITCT}, including the text-to-SQL task.  
For example, Figure~\ref{intro} shows that the baseline model can learn the correct linking among \texttt{date}, \texttt{id} and \texttt{transcript} between the question and schema, but fail to identify that \texttt{id} should also be included in the \texttt{SELECT} clause.
On the other hand, with the help of the dependency tree, \texttt{date} and \texttt{id} are close to each other and thus should appear in the \texttt{SELECT} clause simultaneously.
However, almost all available approaches treat the language question as a sequence, and syntactic information is ignored in neural network-based text-to-SQL models.

\paragraph{\textit{\textbf{Entangled Edge Embedding.}}}
The question-schema interaction graph pre-defines a series of edges, and models them as learnable embeddings.
These embeddings should be diverse by nature because each of them represents a different type of relations and has a different meaning.
Previous work \cite{Brock2019LargeSG,Zhang2020SelfOrthogonalityMA} has proved that the learnable embeddings are easy to be entangled and do not satisfy the diversity objective.

In this paper, we propose \textbf{S$^2$SQL}, injecting \textbf{S}yntax to question-\textbf{S}chema graph encoder for Text-to-\textbf{SQL} parsers. S$^2$SQL models the syntactic labels from a syntactic dependency tree as additional edge embeddings. Motivated by the belief that if the structure of input can be reliably obtained and is a central feature of a task, models that explicitly exploit the structure can benefit. In this paper, we investigate and prove that properly introducing syntactic information into text-to-SQL can further improve the performance, and we provide a detailed analysis on why and how the proposed model works.
Built on that, we propose a decoupling constraint to encourage the model to learn a diverse set of relation embeddings, which further enhances the network's performance. 
We evaluate our proposed model on the challenging text-to-SQL benchmark Spider \cite{YuZYYWLMLYRZR18} and robustness setting Spider-Syn \cite{spider-syn}, and demonstrate that S$^2$SQL outperforms other graph-based models consistently when augmented with different pre-training models.
In brief, the contributions of our work are three-fold:
\begin{itemize}
    \item We investigate the importance of syntax in text-to-SQL and propose a novel and strong encoder for cross-domain text-to-SQL, namely S$^2$SQL.
    \item To induce the diverse edge embedding learning, we introduce the decoupling constraint, which further improves the performance.
    \item The empirical results show that our approach outperforms all the existing models on the challenging Spider and Spider-Syn benchmark. 
\end{itemize}

\section{The Proposed Method}

\subsection{Problem Definition}
Given a natural language question $\mathcal{Q} = \left\{q_{i}\right\}_{i=1}^{|\mathcal{Q}|}$ and a schema $\mathcal{S}=\langle\mathcal{C}, \mathcal{T}\rangle$ consisting of columns $\mathcal{C} = \left\{c_{1}^{t_{1}}, c_{2}^{t_{1}}, \cdots, c_{1}^{t_{2}}, c_{2}^{t_{2}}, \cdots\right\}$ and tables $\mathcal{T} =\left\{t_{i}\right\}_{i=1}^{|\mathcal{T}|}$, text-to-SQL aims to generate the SQL query $y$ for the question sentence. 
The \textit{de facto} method for text-to-SQL employs an encoder-decoder architecture.
In this paper we focus on improving the encoder part.
For a detailed description of the decoder, please refer to the work of \cite{WangSLPR20,lgesql}.

\subsection{Question-Schema Interaction Graph}
The joint input questions and schema items can be viewed as a graph $\mathcal{G}=(\mathcal{V}, \mathcal{R})$, where $\mathcal{V} = \mathcal{Q} \cup \mathcal{T} \cup \mathcal{C}$ are nodes of three types $\{\mathcal{Q}, \mathcal{T}, \mathcal{C}\}$. 
The initial node embeddings matrix $\mathbf{X} \in \mathbb{R}^{\left|V^{|\mathbf{\mathcal{Q}}|+|\mathbf{\mathcal{T}}| + |\mathbf{\mathcal{C}}|}\right| \times d}$ is obtained by flattening all question tokens and schema items into a sequence:
$\texttt{[CLS]} q_{1} q_{2} \cdots q_{|Q|}\texttt{[SEP]} t_{10} t_{1} c_{10}^{t_{1}} c_{1}^{t_{1}} c_{20}^{t_{1}} c_{2}^{t_{1}} \cdots \\ t_{20} t_{2} c_{10}^{t_{2}} c_{1}^{t_{2}} c_{20}^{t_{2}} c_{2}^{t_{2}} \cdots \texttt{[SEP]} $.
The type information $t_{i 0}$ or $c_{j 0}^{t_{i}}$ is inserted before each schema item.
The edge $\mathcal{R}=\{R\}_{i=1, j=1}^{|X|,|X|}$ represents the known relation between two elements in the input nodes.
The RGAT (relational graph attention transformers) \cite{Shaw2018SelfAttentionWR,WangSLPR20,lgesql} models the graph $\mathcal{G}$ and computes the output representation $\mathbf{z}$ by:
\begin{equation}
\resizebox{.8\hsize}{!}{$
\begin{aligned}
e_{i j}^{(h)}&=\frac{\mathbf{x}_{i} \mathbf{W}_{q}^{(h)}\left(\mathbf{x}_{j} \mathbf{W}_{k}^{(h)}+\mathbf{r}_{i j}^{K}\right)^{\top}}{\sqrt{d_{z} / H}}, \\
\alpha_{i j}^{(h)}&=\operatorname{softmax}\left\{e_{i j}^{(h)}\right\}, \\
\mathbf{z}_{i}^{(h)}&=\sum_{v_{j}^{n} \in \mathcal{N}_{i}^{n}} \alpha_{i j}^{(h)}\left(\mathbf{x}_{j} \mathbf{W}_{v}^{(h)}+\mathbf{r}_{i j}^{V}\right),
\end{aligned}
$}
\end{equation}

where matrices $\mathbf{W}_{q}, \mathbf{W}_{k}, \mathbf{W}_{v}$ are trainable parameters in self-attention, and $\mathcal{N}_{i}^{n}$ is the receptive field of node $v_i^n$.

\paragraph{Injecting Syntax} The previous work mainly focuses on using linking structure and schema structure in the encoder \cite{WangSLPR20}, in which the structure of the question is ignored.
We proposed an effective approach to integrate syntactic dependency information\footnote{We use SpaCy toolkit to construct syntactic information: https://spacy.io/.} into the graph.
A straightforward idea is to treat all dependent types directly as a new edge type.
However, the dependency parser will return 55 different dependency types. Such a large number of edge types will significantly increase the number of relational embedding parameters in S$^2$SQL, leading to over-fitting. 
In order to address this, similar to \cite{Vashishth2018DatingDU}, we induct dependency types into three abstract relations, \texttt{Forward}, \texttt{Backward} and \texttt{NONE}.
In addition, in order to ensure the simplicity of edge embedding, we only consider the first-order relationship. By stacking multi-layer transformers, the model implicitly captures the multi-order relationship without deliberate construction.
Specifically, we compute the distance $\boldsymbol{D}(v_{i}, v_{j})$ between any two tokens $v_i$ and $v_j$ from the question. This distance is set to the first-order distance between $v_i$ and $v_j$ if they have the aforementioned dependency types, and 0 otherwise.
Based on this first-order distance $\boldsymbol{D}$, we model the syntactic relation ${R}_{i j}^{\mathrm{question}}$ between tokens $v_i$ and $v_j$ by one of the three previously defined abstract types:
\begin{equation}
    \resizebox{.85\hsize}{!}{$
    \mathcal{R}_{i j}^{\mathrm{question}}=\left\{\begin{array}{cl}
    \texttt{Forward}, & \text {if } \boldsymbol{D}(v_i, v_j) = 1 \\
    \texttt{Backward}, & \text {if } \boldsymbol{D}(v_j, v_i) = 1 \\
    \texttt{NONE}, & \text {otherwise.}
    \end{array}\right.
    $}
\end{equation}

Overall, as shown in Figure~\ref{fig2}, S$^2$SQL models three structures in the graph $\mathcal{G}$:
\begin{itemize}[leftmargin=0.6cm]
    \item \textbf{Question Structure $\mathcal{R}^{\mathrm{question}}$}: relations that represent syntactic dependency between two question tokens. 
    \item \textbf{Linking Structure $\mathcal{R}^{\mathrm{linking}}$}: relations that align entity in question to the corresponding schema columns or tables. \cite{WangSLPR20}
    \item \textbf{Schema Structure $\mathcal{R}^{\mathrm{schema}}$}: relations within a database schema, \eg, \textit{foreign key}.
\end{itemize}
The detailed structure construction could be found in Appendix \ref{structure}.

\paragraph{Decoupling Constraint.} There are $k$ known edges in $\mathcal{R}$ and each is represented as a relation embedding.
Intuitively, these edge embedding $\mathbf{r} =  [\mathbf{r}_1, \mathbf{r}_2,..., \mathbf{r}_k ]$ should be diverse because they have different semantic meanings.
To avoid the potential risk of coupling edge embedding $\mathbf{r}$ during optimization, we introduce the orthogonality condition \cite{Brock2019LargeSG} to $\mathbf{r}$:
\begin{equation}
\resizebox{.6\hsize}{!}{$
\mathcal{L}(\mathbf{r})=\left\|\mathbf{r}^{\top} \mathbf{r} \odot(\mathbf{1}-I)\right\|_{\mathrm{F}}^{2},
$}
\end{equation}
where $\mathbf{1}$ denotes a matrix with all elements being set to 1 and $I$ is the identity matrix.

\section{Experiments}

\subsection{Experiment Setup}

\paragraph{Datasets and Evaluation Metrics.}
We conduct experiments on Spider \citep{YuZYYWLMLYRZR18} and Spider-Syn \citep{spider-syn}. Spider is a large-scale, complex, and cross-domain text-to-SQL benchmark. 
Spider-Syn is derived from Spider, by replacing their schema-related words with manually selected synonyms that reflect real-world question paraphrases.
For evaluation, we followed the official evaluation to report exact match accuracy. 

\subsection{Implementation Details.}
\label{details}
We utilize PyTorch \cite{Paszke2019PyTorchAI} to implement our proposed model. 
During pre-processing, the input of questions, column names, and table names are tokenized and lemmatized with the Standford Nature Language Processing toolkit. 
For a fair comparison with baselines, we configure it with the same set of hyper-parameters, \eg, stacking 8 self-attention layers, setting dropout to 0.1. The position-wise feed-forward network has an inner layer dimension of 1024. Inside the decoder, we use rule embeddings of size 128, node type embeddings of size 64, and a hidden size of 512 inside the LSTM with a dropout of 0.21.

\begin{table}[t]  
    \centering
    \scalebox{0.8}{
    \begin{tabular}{lcc}  
    \toprule
    \textbf{Model}& \textbf{Dev.} & \textbf{Test}\\ 
    \midrule
    Global-GNN \cite{BoginGB19} & 52.7 & 47.4 \\
    Eidt-SQL \cite{ZhangYESXLSXSR19}& 57.6 & 53.4 \\
    Bertand-DR \cite{Kelkar2020BertrandDRIT} & 57.9 & 54.6 \\
    IRNet v2 \cite{guo-etal-2019-towards} & 63.9 & 55.0 \\
    BRIDGE \cite{lin-etal-2020-bridging}& 70.0 & 65.0 \\
    RYANSQL \cite{Choi2020RYANSQLRA} & 70.6 & 60.6 \\
    RATSQL + BERT \cite{WangSLPR20}& 69.7 & 65.6 \\
    ShadowGNN + RoBERTa \cite{Chen2021ShadowGNNGP} & 72.3 & 66.1 \\
    \midrule
    RAT + RoBERTa \cite{WangSLPR20}  & 69.7 & 64.3 \\ 
    S$^2$SQL + RoBERTa & 71.4 & \textbf{67.1}  \\
    \quad w/o DC & 70.9 & - \\
    \midrule
    LGESQL + ELECTRA \cite{lgesql}  & 75.1 & 72.0 \\ 
    S$^2$SQL + ELECTRA & \textbf{76.4}  & \textbf{72.1}  \\
    \quad w/o DC & 75.8 & - \\
    \bottomrule
    \end{tabular}  
    }
    \caption{The exact match accuracy on the Spider dev and test set. 
    $-$ indicates that the test set performance cannot be obtained due to the number of submission limit.}
    \label{tab1}
\end{table}  

\begin{figure}
    \small
    \centering
    \includegraphics[width=0.45\textwidth]{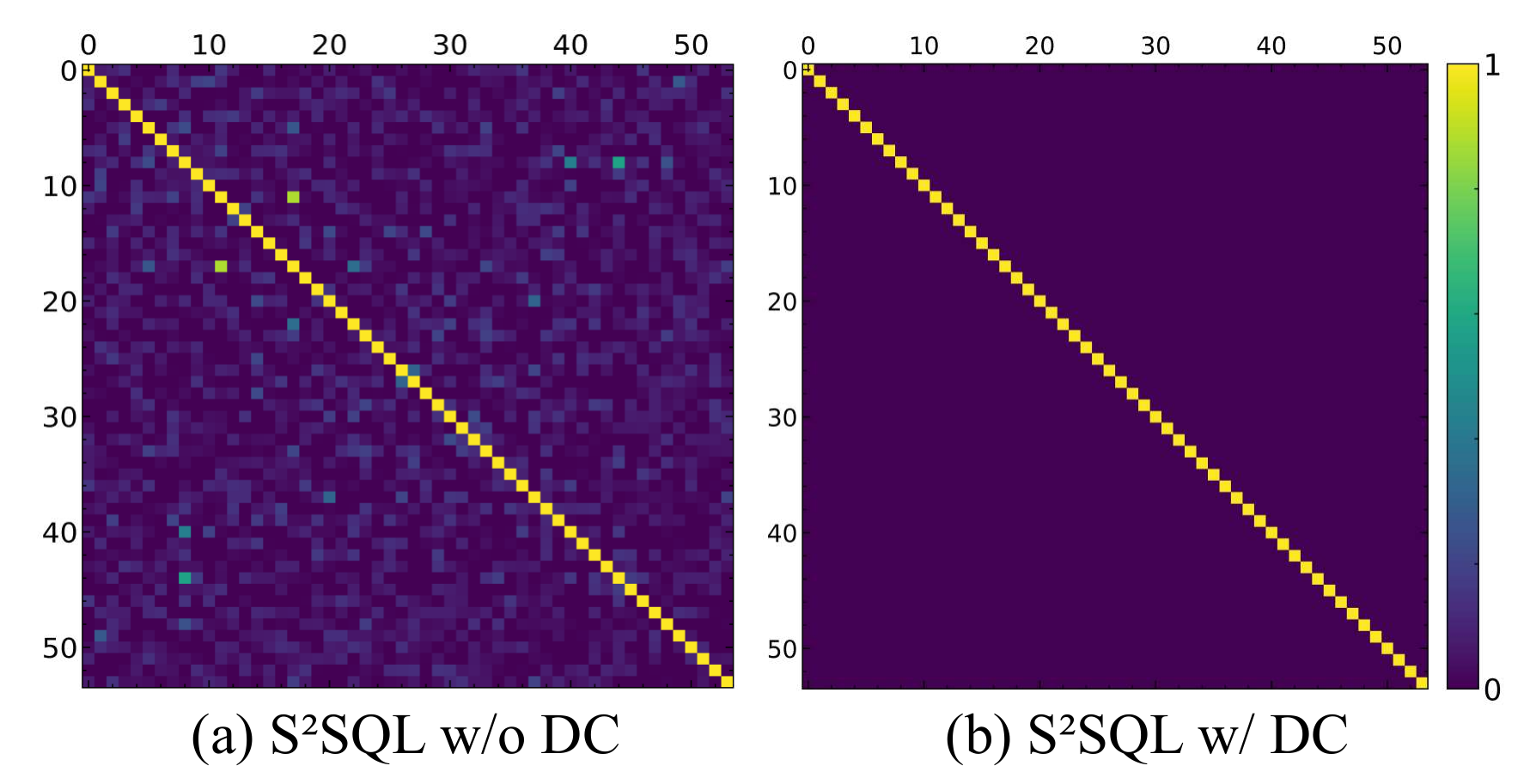}
    \caption{The similarity matrix of different relation embeddings with and without DC.
    The lighter the color, the higher the similarity (entangled embeddings).}
    \label{DC}
\end{figure}

\subsection{Baseline Models.}
\label{baselines}
We conduct experiments on Spider and Spider-Syn and compare our method with several baselines including:
\begin{itemize}
    \item \textbf{RYANSQL} \cite{Choi2020RYANSQLRA} is a sketch-based slot filling approach which is proposed to synthesize each SELECT statement for its corresponding position.
    \item \textbf{RATSQL} \cite{WangSLPR20} is a relation aware schema encoding model in whuich the question-schema interaction graph is built by n-gram patterns.
    \item \textbf{ShadowGNN} \cite{Chen2021ShadowGNNGP} processes schemas at abstract and semantic levels with domain-independent representations.
    \item \textbf{BRIDGE} \cite{lin-etal-2020-bridging} represents the question and schema in a tagged sequence where a subset of the fields are augmented with cell values mentioned in the question.
    \item \textbf{LGESQL} \cite{lgesql} a line graph enhanced Text-to-SQL model to mine the underlying relational features without constructing metapaths. It was the SOTA model in the Spider leaderboard before ours.
\end{itemize}

\subsection{Results and Analyses}

\paragraph{Overall Performance.} 
We first compare S$^2$SQL with other state-of-the-art models on Spider. As shown in Table\ref{tab1}, we can see that S$^2$SQL outperforms all existing models.
Remarkably, the accuracy of S$^2$SQL + RoBERTa on the hidden test set is 67.1\%, which is 2.8\% higher than the strong baseline RAT + RoBERTa. 
Similarity, the accuracy of the SOTA model LGESQL + ELECTRA is 72.0\% on the hidden test set, and 75.1\% on the development set, while S$^2$SQL + ELECTRA can reach 72.1\% test and 76.4 development accuracy.
Table \ref{tab2} shows results on the development set for RAT and S$^2$SQL with Table-based pre-training models. We can see that S$^2$SQL outperforms RAT consistently when augmented with different pre-training models, including RoBERTa \cite{Liu2019RoBERTaAR}, GraPPa \cite{Yu2020GraPPaGP} and GAP \cite{Shi2020LearningCR}.
In addition, as shown in Table \ref{tab3}, S$^2$SQL demonstrates improvement on the robustness dataset.

\begin{table}[t]
    \centering
    \scalebox{0.9}{
    \begin{tabular}{lc}
    \toprule
    \textbf{Model}& \textbf{Dev.} \\ 
    \midrule
    RAT + GraPPa \cite{Yu2020GraPPaGP} $\dagger$ & 71.5  \\
    S$^2$SQL + GraPPa & \textbf{73.4} \\
    \midrule
    RAT + GAP \cite{Shi2020LearningCR} $\dagger$ & 71.8 \\
    S$^2$SQL + GAP & \textbf{72.7} \\
    \bottomrule
    \end{tabular}}
    \caption{Comparison on S$^2$SQL under the different table-based pre-training models on Spider Dev set.}
    \label{tab2}
\end{table}

\begin{table}[t]
    \centering
    \scalebox{0.9}{
    \begin{tabular}{lc}
    \toprule
    \textbf{Model}& \textbf{Acc.} \\ 
    \midrule
    Global-GNN \cite{BoginGB19} & 23.6 \\
    IRNet \cite{guo-etal-2019-towards} & 28.4 \\
    RATSQL \cite{WangSLPR20} & 33.6 \\
    RATSQL + BERT \cite{WangSLPR20} & 48.2 \\
    RATSQL + Grappa \cite{WangSLPR20} & 49.1 \\
    S$^2$SQL + Grappa & \textbf{51.4} \\
    \bottomrule
    \end{tabular}}
    \caption{The accuracy on the Spider-Syn dataset.}
    \label{tab3}
    \vspace{-0.3cm}
\end{table}

\begin{table*}[!]
\centering
\resizebox{1\hsize}{!}{
\begin{tabular}{ll}
\toprule
 Question & \textit{List the name and tonnage in alphabetical descending order for the names.
}\\
 Baseline & SELECT name, tonnage FROM ship ORDER BY \bad{tonnage} DESC \\
 S$^{2}$SQL & SELECT name, tonnage FROM ship ORDER BY \good{name} DESC \\
 Gold & SELECT name, tonnage FROM ship ORDER BY name DESC \\ 
 Syntax & (name, tonnage, CONJ), (order, names, NMOD) \\
 \midrule
 Question & \textit{What is the total population and average area of countries in the continent of North America whose area is bigger than 3000 ?}\\
 Baseline & SELECT sum(population) ,  avg(surface\_area) FROM country where \bad{surface\_area}  =  “North America" and surface\_area  >  3000 \\
 S$^{2}$SQL & SELECT sum(population) ,  avg(surface\_area) FROM country where \good{continent}  =  “North America" and surface\_area  >  3000 \\
 Gold & SELECT sum(population) ,  avg(surface\_area) FROM country where continent  =  “North America" and surface\_area  >  3000 \\
 Syntax & (continent, America, NMOD) \\
 \midrule
 Question & \textit{Show the date of the transcript which shows the least number of results, also list the id.}\\
 Baseline & SELECT transcript\_date FROM Transcript\_Contexts AS T1 JOIN … \\
 S$^{2}$SQL & SELECT transcript\_date, \good{transcript\_id} FROM Transcript\_Contexts AS T1 JOIN … \\
 Gold & SELECT transcript\_date, transcript\_id FROM Transcript\_Contexts AS T1 JOIN … \\
 Syntax & (show, list, DEP) (show, date, OBJ) (list, id, OBJ)\\ 
 \bottomrule
\end{tabular}
}
\vspace{-0.2cm}
\caption{Case study: some comparisons with baseline (LGESQL) show that S$^2$SQL can generate more accurate SQL, where syntax column represents useful syntactic information in the generation of S$^2$SQL.}
\vspace{-0.5cm}
\label{tab:case}
\end{table*}

\paragraph{Ablation Study.}
The last row of Table~\ref{tab1} shows that removing the decoupling constraint causes a $~$0.5\% performance drop on the development set.
This implies that decoupling entangled embeddings helps to improve the performance.
To examine the impact of the decoupling constraint, we visualize the cosine similarity between any two relation embeddings.
As shown in Figure~\ref{DC}, we observe that the decoupling constraint eliminates the entangling phenomenon (darker colors) and produces a more diverse set of embeddings.

\subsection{Qualitative Analysis.}
\label{analysis}
In Table \ref{tab:case}, we compare the SQL queries generated by our S$^2$SQL model with those created by the baseline model LGESQL. We notice that S$^2$SQL performs better than the baseline system, especially in the case of question understanding that depends on syntax structure. For example, in the first case where the \texttt{order} and \texttt{name} have \texttt{NMOD} relation, baseline fails to   
For example, in the first example, both \texttt{name} and \texttt{tonnage} can be linked correctly, but the baseline fails to capture the structure present in \texttt{name} and \texttt{order}, resulting in a generation error, while S$^2$SQL predicts the result well.

\subsection{About Syntactic Parser.}
In our experiments, we use the SpaCy tool as a syntactic parser. It is important to emphasize that the quality of the SpaCy syntactic parsing has marginal impact on the performance of S2SQL. The following three main reasons are given.
\begin{itemize}
    \item SpaCy is the current SOTA parser tool (95\%+ accuracy on the OntoNotes 5.0 corpus) and has been widely used in various papers introducing syntax, which proves its reliability.
    \item The question in Spider are not extremely complex and can be handled very well.
    \item Even though syntactic parser errors may introduce noise into S2SQL, our proposed inductive syntactic injection method (instead of independent injection) can mitigate the impact of syntactic type errors.
\end{itemize}

\section{Related Work}
\vspace{-0.2cm}
Extensive work has been conducted on improving the encoder and decoder \cite{YinN17,wang2019learning,guo-etal-2019-towards,Choi2020RYANSQLRA,Kelkar2020BertrandDRIT,rubin2020smbop,Hui2021ImprovingTW} as well as table-based pre-training \cite{Yin2020TaBERTPF,Yu2020GraPPaGP,Deng2020StructureGroundedPF,Shi2020LearningCR,wang2021learning}.
Besides, \citet{wang2020meta} proposed a meta-learning based training objective to boost generalization.
\citet{Scholak2021PICARDPI} proposed PICARD, a method for constraining auto-regressive decoders of T5. 
Among the encoder-related work,
\citet{guo-etal-2019-towards} introduced the schema linking module, which aimed to recognize the columns and the tables mentioned in a question.
\citet{lin-etal-2020-bridging} leveraged the database content to augment the schema representation. 
\citet{BoginGB19} employed GNN to derive the representation of the schema structure.
Then, \citet{Chen2021ShadowGNNGP} proposed ShadowGNN to abstract the representation of question and schema with attention. 
Besides, \citet{Hui2021DynamicHR} present a dynamic graph framework that can model contextual information for context-dependent setting.
The most recent approaches \cite{WangSLPR20,lgesql} achieved the best performance through relation-aware transformer. 
Unlike these works, we investigated the impact of the syntactic structures during the encoding stage.

\vspace{-0.2cm}
\section{Conclusion}
\vspace{-0.2cm}
We present syntax-enhanced question-schema graph encoder (S$^2$SQL) that can effectively model syntactic information for text-to-SQL and introduce the decoupling constraint to induce the diverse relation embedding. The proposed model achieves new state-of-the-art performance on the widely used benchmark, Spider and Spider-Syn.

\bibliography{anthology,custom}
\bibliographystyle{acl_natbib}

\newpage

\appendix

\section{Appendix}
\label{sec:appendix}

\subsection{Details of Relation Structure.}
\label{structure}
All structures have been shown in Table \ref{graph}.
a structure (edge) exists from source node $x \in S$ to target node $y \in S$ if the pair fulfills one of the descriptions listed in the Table \ref{graph}, with the corresponding label.

\begin{table*}[!]
\centering
\resizebox{1\hsize}{!}{
\begin{tabular}{llll}
\toprule
Source $x$ & Target $y$ & Type & Description \\
\midrule
Question & Question & \texttt{Forward-Syntax} & y is the target word of x under syntax dependency. \\
Question & Question & \texttt{Backward-Syntax} & y is the source word of x under syntax dependency. \\  
Question & Question & \texttt{None-Syntax} & x and y have no syntactic dependency. \\
\midrule
Column & Column & \texttt{Foreign-Key} & y is the foreign key of x. \\
\midrule
Table & Column & \texttt{Has} & The column y belongs to the table x. \\
Table & Column & \texttt{Primary-Key} & The column y is the primary key of the table x. \\
\midrule
Question & Table & \texttt{None-Linking} & No linking between x and y. \\
Question & Table & \texttt{Partial-Linking} & x is part of y, but the entire question does not contain y. \\
Question & Table & \texttt{Exact-Linking} & x is part of y, and y is a span of the entire question. \\
\midrule
Question & Column & \texttt{None-Linking} & No linking between x and y. \\
Question & Column & \texttt{Partial-Linking} & x is part of y, but the entire question does not contain y. \\
Question & Column & \texttt{Exact-Linking} & x is part of y, and y is a span of the entire question. \\
Question & Column & \texttt{Value-Linking} & x is part of the candidate cell values of column y. \\
 \bottomrule
\end{tabular}
}
\caption{The checklist of all relations structure used in S$^2$SQL. All relations above are asymmetric.}
\label{graph}
\end{table*}

\end{document}